\documentclass[letterpaper, 10 pt, conference]{ieeeconf}
\IEEEoverridecommandlockouts
\usepackage{cite}
\usepackage{amsmath,amssymb,amsfonts}
\usepackage{graphicx}
\usepackage{textcomp}
\usepackage{xcolor}
\usepackage[pdftex]{hyperref}
\usepackage[noabbrev]{cleveref}

\usepackage{algorithm}
\usepackage[noend]{algpseudocode}

\usepackage{caption}
\usepackage{subcaption}

\usepackage{todonotes}
\def\BibTeX{{\rm B\kern-.05em{\sc i\kern-.025em b}\kern-.08em
    T\kern-.1667em\lower.7ex\hbox{E}\kern-.125emX}}

\usepackage[acronym, shortcuts]{glossaries}

\usepackage{booktabs}

\definecolor{bluegreen}{HTML}{4488b2}
\definecolor{redpink}{HTML}{f4858d}
\definecolor{darkblue}{HTML}{4b50fc}

\definecolor{yellowgreen}{HTML}{98ff4d}
\definecolor{rust}{HTML}{a24443}
\definecolor{orange}{HTML}{f2a147}

\newacronym{drl}{DRL}{deep reinforcement learning}
\newacronym{pacmap}{PaCMAP}{Pairwise Controlled Manifold Approximation Projection}
\newacronym{rl}{RL}{reinforcement learning}
\newacronym{sac}{SAC}{Soft Actor-Critic}
\newacronym{ddpg}{DDPG}{Deep Deterministic Policy Gradient}
\newacronym{pca}{PCA}{Principal component analysis}
\newacronym{gan}{GAN}{Generative Adversarial Network}

\begin{document}

\title{Discovering Behavioral Modes in Deep Reinforcement Learning Policies Using Trajectory Clustering in Latent Space}

\author{Sindre Benjamin Remman$^{1}$ and Anastasios M. Lekkas$^{2}$
\thanks{$^{1}$Sindre Benjamin Remman is with the Department of Engineering Cybernetics, 
        Norwegian University of Science and Technology (NTNU), Trondheim, Norway
        {\tt\small sindre.b.remman@ntnu.no}}%
\thanks{$^{2}$Anastasios M. Lekkas is with the Department of Engineering Cybernetics,
        Centre for Autonomous Marine Operations and Systems (AMOS),
        Norwegian University of Science and Technology (NTNU), Trondheim, Norway
        {\tt\small anastasios.lekkas@ntnu.no}}%
}

\maketitle

\begin{abstract}

Understanding the behavior of deep reinforcement learning (DRL) agents is crucial for improving their performance and reliability. However, the complexity of their policies often makes them challenging to understand. In this paper, we introduce a new approach for investigating the behavior modes of DRL policies, which involves utilizing dimensionality reduction and trajectory clustering in the latent space of neural networks. Specifically, we use Pairwise Controlled Manifold Approximation Projection (PaCMAP) for dimensionality reduction and TRACLUS for trajectory clustering to analyze the latent space of a DRL policy trained on the Mountain Car control task. Our methodology helps identify diverse behavior patterns and suboptimal choices by the policy, thus allowing for targeted improvements. We demonstrate how our approach, combined with domain knowledge, can enhance a policy's performance in specific regions of the state space.

\end{abstract}


\section{Introduction}

Recent developments in \ac*{drl} have shown impressive performance in various fields, from video games \cite{badia2020agent57} to robotic control \cite{brunke2022safe}. However, understanding the decision-making process of \ac*{drl} agents remains a significant challenge due to the complexity and opacity of their neural network policies. This lack of interpretability hinders the ability to diagnose and improve agent performance in control tasks, especially in critical applications. To address this problem, we introduce a novel approach to investigate and understand the behaviors of DRL agents by examining their learned policies through unsupervised learning.

We propose a methodology that first involves using \ac*{pacmap} \cite{wang2021understanding} to map the latent space trajectories of a \ac*{drl} policy into a lower-dimensional space. \ac*{pacmap} is a method that aims to preserve the data's local and global structure while reducing the data's dimensionality. After reducing the dimensionality, we apply TRACLUS \cite{lee2007trajectory}, a trajectory clustering algorithm, which helps us to discern distinct \emph{behavior modes} of the agent's policy. Behavior modes, in this context, are defined as consistent approaches or tactics to specific situations or states that the agent encounters.

To capture comprehensive behavioral patterns that extend over time, we cluster trajectories instead of individual states or actions. Each trajectory includes a set of points, each point corresponding to a time step in one episode of a \ac*{drl} agent's operation in the environment. We generate the latent space representation of this trajectory by inputting the state into the neural network policy at each time step. We then extract the latent space representation of the data from the second-to-last layer of the neural network.

By visualizing the clusters in a plot, our approach not only provides a visual representation of the policy's strategy but also enables us to reveal suboptimal behavior patterns. Our focus in this paper is the Mountain Car problem, a classic control task with straightforward mechanics and a two-dimensional state space, simplifying visualizations of the states. By analyzing the critical decision regions in the state space, such as the choice between direct uphill attempts toward the goal or accumulating mechanical energy, we detect nuances in the agent's behavior that are not immediately apparent. Understanding these nuances allows us to make targeted improvements to the agent's policy, leading to measurable performance enhancements. 

In \cite{mannor2004dynamic}, a somewhat related approach to our proposed methodology is employed, which uses the \emph{options} reinforcement learning framework \cite{sutton1999between}. The authors of \cite{mannor2004dynamic} use an algorithm to divide the state space into several clusters, learn options for moving between the clusters, and finally use these options as \emph{macro-actions} for solving the environments using Q-learning \cite{watkins1992q}. Although \cite{mannor2004dynamic} uses a very different methodology than the one presented in here, the resulting clusters could be interpreted similarly. 

Another related approach can be seen in \cite{mukherjee2019clustergan}, where the proposed approach, ClusterGAN, leverages the latent space of a \ac*{gan} for clustering. Notably, the approach in \cite{mukherjee2019clustergan} enables \ac*{gan}s to perform smooth latent space interpolations between categories, regardless of the discriminator's training on such data representations. Although \cite{mukherjee2019clustergan} does not involve \ac*{drl} or trajectory clustering, they still, similar to our proposed methodology, cluster in the latent space of a neural network.

Although different from our proposed methodology, work has also been done to explain and interpret \ac*{drl} policies. The authors of \cite{gjaerum2023model} use model tree methods as a surrogate for a policy controlling an autonomous surface vehicle for docking at a harbor and use this surrogate for approximating policy feature attributions. The authors of \cite{guo2021edge} propose a self-explainable model that predicts the agent's final rewards from the game episodes and extracts time step importance within the episodes as strategy-level explanations for the agent. The authors of \cite{he2021explainable} propose an explainable deep reinforcement learning path planner for flying a quadcopter through an unknown environment. 

While the existing studies offer valuable frameworks and insights, our method provides a novel approach that aids the understanding of an agent's decision-making over time. Specifically, this paper has the following key contributions:

\begin{itemize}
    \item We employ \ac*{pacmap}, a state-of-the-art dimension reduction method, to reduce the dimensionality of the latent space of a \ac*{drl} policy.
    \item Leveraging \ac*{pacmap}'s representation of the latent space, we employ TRACLUS to cluster trajectories within the latent space of the policy.
    \item We interpret the resulting clusters generated by the combination of \ac*{pacmap} and TRACLUS and use them for behavior classification of the \ac*{drl} Policy across different parts of the environment's state space. 
    \item We detect areas of the state space where the policy is performing poorly by combining domain knowledge with the behavior clusters generated by the proposed methodology. We subsequently show how the policy's behavior within these areas of the state space could be improved.
\end{itemize}

To our knowledge, this research represents the first application of trajectory clustering within the latent spaces of DRL policies to detect their behavior modes.

\section{Preliminaries}\label{sec:preliminaries}

\subsection{TRACLUS}

TRACLUS, introduced by Lee, Han, and Whang in 2007 \cite{lee2007trajectory}, is a trajectory clustering algorithm that identifies common sub-trajectories across sets of trajectories. The algorithm's applications span hurricane tracking, wildlife tracking \cite{lee2007trajectory}, visualizing singing styles \cite{lin2014visualising}, and transportation \cite{mustafa2021gtraclus}.

Built on a \emph{partition-and-group} framework \cite{lee2007trajectory}, TRACLUS comprises two main phases: 
\begin{enumerate}
    \item \textbf{Partitioning Phase:} Each trajectory is divided into line segments based on \emph{characteristic points}, identified by the sub-algorithm \emph{Approximate Trajectory Partitioning}. These points indicate rapid changes in a trajectory's behavior.
    \item \textbf{Grouping Phase:} Segments are clustered, irrespective of their originating trajectory. Clustering employs a modified DBSCAN \cite{ester1996density}, with distances determined by three functions: \emph{perpendicular distance}, \emph{parallel distance}, and \emph{angle distance} \cite{lee2007trajectory, chen2003noisy}.
\end{enumerate}

The algorithm determines a segment's neighbors using the distance functions mentioned above. Two line segments are neighbors if the distance between them is below a certain threshold, $\epsilon$. Using this definition of neighboring line segments, TRACLUS assigns every line segment to be one of three different types:
\begin{itemize}
    \item \emph{Core line segments}:
    \begin{itemize}
        \item These line segments have more neighbors than a certain threshold parameter, \emph{MinLns}. Core line segments are core to each cluster, and every cluster has to have at least one core line segment.
    \end{itemize}
    \item \emph{Border line segments}:
    \begin{itemize}
        \item These line segments are neighbors with a core line segment but have less than \emph{MinLns} neighbors. Border line segments define the border of the clusters. 
    \end{itemize}
    \item \emph{Noise line segments}:
    \begin{itemize}
        \item TRACLUS defines every line segment, not a core line segment or a border line segment, to be noise. 
    \end{itemize}
\end{itemize}


\subsection{PaCMAP}\label{subsec:pacmap}

Introduced by Wang, Huang, Rudin, and Shaposhnik in 2020 \cite{wang2021understanding}, \ac*{pacmap} aims to reduce the dimensionality of a dataset while preserving both the global and local structure in the data. By examining previous dimension-reduction methods, the authors found that there are specific properties of the loss function (and, to a lesser extent, the weights) that are very important for preserving local structure in the data when embedding it in a lower-dimensional space. In addition, they found that the choice of graph components is crucial for preserving the data's global structure.

\ac*{pacmap} uses three different kinds of graph components in its loss function: nearest neighbor edges (denoted as neighbor edges), mid-near edges (MN edges), and repulsion edges (FP edges). Neighbor edges are edges between a pair of near neighbors, where each data point $i$ is paired with its nearest $n_{NB}$ neighbors defined by the scaled distance in the higher-dimensional space, $$d^{2,\text{select}}_{i,j} = \frac{||\mathbf{x}_i - \mathbf{x}_j||^2}{\sigma_i \sigma_j},$$ where $\sigma_i$ is the average Euclidean distance between $i$ and its nearest fourth to sixth neighbors in the higher-dimensional space, and $x_i$ is $i$'s data in the higher-dimensional space. MN edges are found by, for each $i$, sampling six observations, selecting the second closest observation to $i$ out of these six, and pairing them. Finally, FP edges are chosen by sampling non-neighbors. The algorithm's three most important parameters each correspond to one of these types of graph components. $n_{NB}$ selects how many nearest neighbors the algorithm should use; $MN_{ratio}$ sets the ratio of MN edges to $n_{NB}$, with the default value being $0.5$; and $FP_{ratio}$ is the ratio of FP edges to $n_{NB}$, where the default value is $2.0$.

\ac*{pacmap}'s loss function is as follows:

\begin{align*}
    Loss^{PaCMAP} =\text{ }&w_{NB} \sum_{\text{i, j are neighbors}} 
    \frac{\tilde{d}_{ij}}{10+\tilde{d}_{ij}}\\ + &w_{MN} \sum_{\text{i, j are mid-near pairs}} 
    \frac{\tilde{d}_{ij}}{10000+\tilde{d}_{ij}}\\ + &w_{FP} \sum_{\text{i, j are FP pairs}}
    \frac{1}{1+\tilde{d}_{ij}},
\end{align*}
where $\tilde{d}_{ij}=||\mathbf{y}_i - \mathbf{y}_j||^2$. $w_{NB}$, $w_{MN}$, and $w_{FP}$ are the weights for the three graph components and are changed according to a pre-determined scheme throughout the algorithm's iterations. This loss function is optimized over several iterations using an optimizer such as Adam \cite{wang2021understanding}.

\section{Methodology}\label{sec:methodology}

\begin{figure}
    \centering
    \includegraphics[width=.7\linewidth]{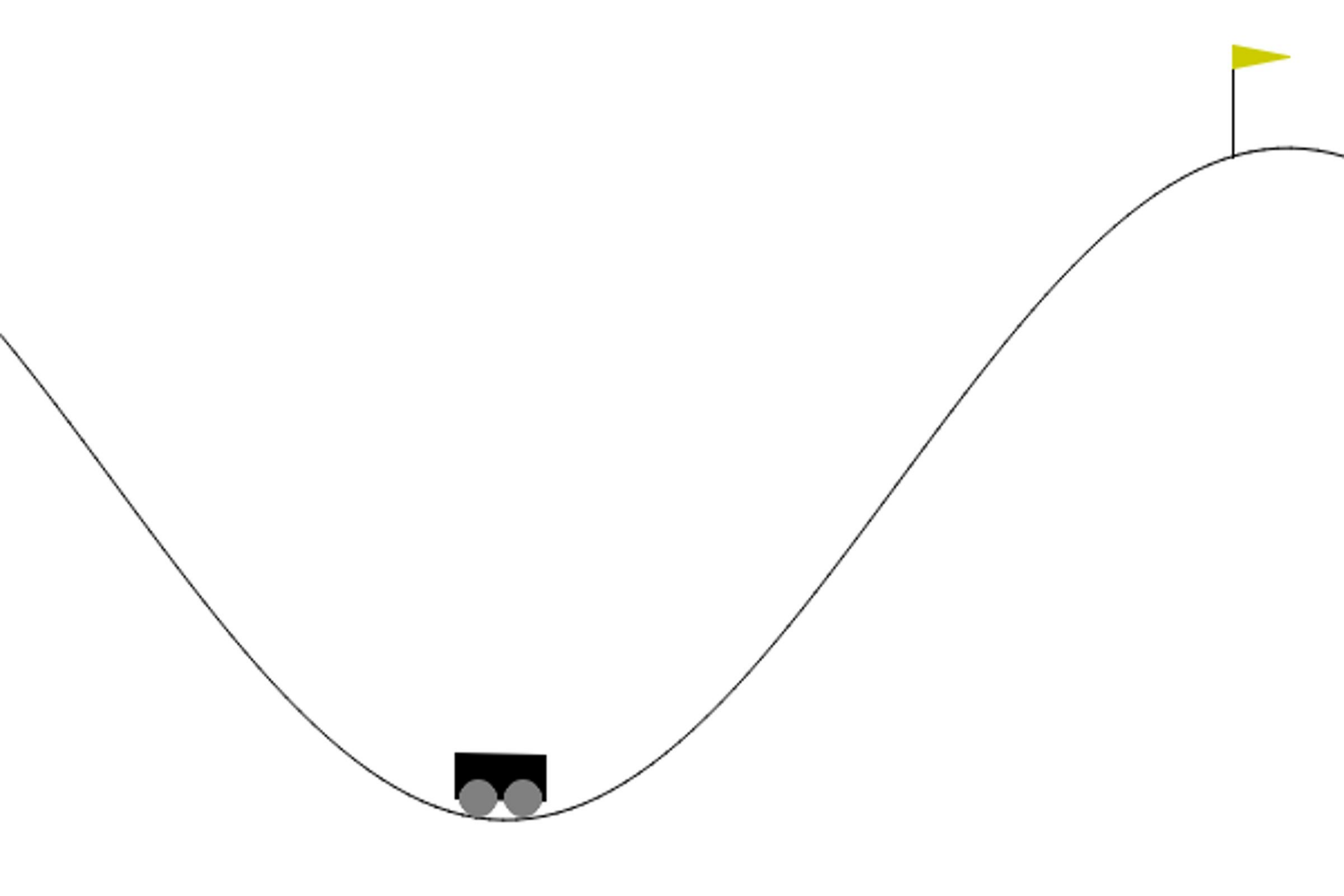}
    \caption{MountainCarContinuous-v0 environment}
    \label{fig:mountaincar}
    \vspace{-0.4cm}
\end{figure}

We apply our methodology to a modified MountainCarContinuous-v0 environment from OpenAI Gymnasium \cite{towers_gymnasium_2023}, seen in \cref*{fig:mountaincar}. In this environment, a car starts at the bottom of a valley, and the only action available is to accelerate the car in either direction. The positive direction is to the right for the position, velocity, and acceleration. The MDP is deterministic, but the car's starting position is stochastic. The objective is to accelerate the car strategically to reach the goal state at the top of the right hill, at car position $= 0.45$. The environment's original reward function is

\begin{equation}\label{eq:original_reward_function}
    r_t=
\begin{cases}
100, & \text{for } pos_t \geq 0.45\\
-a_t^2, & \text{otherwise}
\end{cases},
\end{equation}

where $r_t$, $pos_t$, and $a_t$ are the reward, position of the cart, and the action applied to the environment, respectively, at time step $t$. For our modified environment, we changed the reward function to

\begin{equation}\label{eq:modified_reward_function}
    r_t=
\begin{cases}
100, & \text{for } pos_t \geq 0.45\\
-1, & \text{otherwise}
\end{cases},
\end{equation}

We made this change to make it more intuitive to understand what the type of behavior at each point in the state space is and to identify possible anomalies in the agent's behavior patterns more easily. With the proposed reward function \labelcref*{eq:modified_reward_function}, the agent aims to reach the goal in the least amount of time, which generally can be seen from a simple plot of the agent's trajectories in the state space. This is because shorter episodes now always correspond to a higher reward for the agent.

\subsection{Training the policy and generating the data set}\label{subsec:meth_training_policies}

We trained the policy using the Stable-Baselines3 \cite{stable-baselines3} implementation of the \ac*{sac} algorithm \cite{haarnoja2018soft}. From now on, we refer to this policy as the \emph{MC Policy}. This policy was trained from a randomly initialized neural network for 100000 time steps. We saved the policy every 1000 time steps and selected the best-performing policy. We tune and choose the hyperparameters using Optuna \cite{optuna_2019}.

To get diverse trajectories that represent the whole state space, we select 100 different initial states and run the fully-trained policy from this until the car reaches the goal. We choose these 100 initial states from a uniform grid of the minimum and maximum state values: $cart_{position} \in [-1.25, 0.5]$ and $cart_{velocity} \in [-0.07, 0.07]$. We generate the latent space representations of the data by sending the states through all layers but the last layer of the policy's neural network. The second to last layer has 64 nodes, which means the dimensionality of the data's latent space representation is also 64. We gather these representations into a data set, where the columns are the output of each of the 64 nodes, and the rows correspond to each data point generated by running the policy from the 100 initial states. \ac*{sac} uses a stochastic policy, but to get deterministic results, we use the mean actions chosen by the policy when generating the data. 


\subsection{Dimensionality reduction}\label{subsec:dim_reduction}

Before we cluster the trajectories in latent space, we reduce the dimensionality of the data's latent space representation. To determine the number of dimensions we should reduce the latent space representation to, we use a heuristic based on \ac*{pca} \cite{mackiewicz1993principal}. We calculate the principal components of the latent space data set and then calculate how much of the variance of the data is explained by each of the principal components. We set a threshold of how much of the variance we want to be explained by the principal components. We use this as a heuristic for determining the number of dimensions to reduce the latent space representation. By trial-and-error, we set the explained variance threshold to $99.9\%$ and see that the 13 first principal components explain approximately over $99.9\%$ percentage of the data set's variance. Hence, we determined we should reduce the dimensionality of the latent space from 64 to 13.

We employ \ac*{pacmap} to reduce the dimensionality of the data's latent space representation while preserving local and global structure. The reasons for reducing the latent space data's dimension are three-fold:

\begin{itemize}
    \item Clustering using distance functions generally works best in lower dimensional spaces because of the \emph{curse of dimensionality} \cite{BellmanDynamicProgramming, beyer1999nearest}. 
    \item Several of the columns of the latent space data sets are either constant or highly correlated with the other columns, indicating an over-dimensioned policy.
    \item \ac*{pacmap} helps structure the data into a format we show empirically in \cref*{sec:results_discussion} helps TRACLUS form more meaningful clusters when we take the data back into state space. 
\end{itemize}

We observe that of the three parameters discussed in \cref*{subsec:pacmap}, the $n_{NB}$ parameter makes the most significant difference in the resulting embedding for our data. To select the value of this parameter, we create two-dimensional embeddings for many different values for $n_{NB}$ and see which embedding most clearly separates the data while keeping the trajectories structured. For the other parameters, we use the default values. 

\subsection{Trajectory clustering}

The latent space representation of the data has now been embedded down to 13 dimensions. Next, we use TRACLUS to cluster the sub-trajectories contained by each trajectory in the reduced space. We implement TRACLUS as described in \cite{lee2007trajectory}. The algorithm has two critical parameters, $\epsilon$ and \emph{MinLns}. We determine $\epsilon$ using the same heuristic as the original authors: by finding the $\epsilon$ which minimizes the data's entropy, where the entropy is defined by

\begin{align}
& H(X) = \sum_{i=1}^n p(x_i) \log_2 \frac{1}{p(x_i)} = - \sum_{i=1}^n p(x_i) \log_2 p(x_i) \\
    & \text{where } p(x_i) = \frac{|N_\epsilon(x_i)|}{\sum_{j=1}^n |N_\epsilon (x_j)} \text{, } n=num_{ln} \nonumber \\
&\text{and } N_\epsilon (L_i) = \{L_j \in \mathcal{D} | dist(L_i, L_j) \leq \epsilon\}. \nonumber
\end{align}

In this entropy definition, $|N_\epsilon (L_i)|$ is the number of neighbors of line segment $L_i$, and $num_{ln}$ is the number of line segments generated by the \emph{Approximate Trajectory Partitioning} algorithm from \cite{lee2007trajectory}. We also use a similar heuristic as \cite{lee2007trajectory} to determine \emph{MinLns}. We set this parameter to be the average number of neighbors each line segment has within the $\epsilon$, which minimizes the entropy, rounded up: $\lceil avg_{|N_\epsilon(L)|} \rceil$. 

One problem with using TRACLUS for our purposes is that many line segments will be classified as \emph{noise}. The algorithm classifies a line segment as noise if it is neither a core nor a border line segment. For our data, using the parameters defined using the heuristics above gives us clusters that illustrate well the general trends and behavior in the data. Still, many line segments are defined as noise and are, therefore, not assigned to any cluster. 
We take all the line segments assigned as noise and greedily assign them to one of the clusters based on which cluster has the $n$ closest line segments to the current noise line segment. This is iteratively done with the new clusters but still with the same line segments initially assigned as noise until the clusters have converged. This procedure is shown in \Cref*{algo:assign_noise_to_clusters}.

\begin{algorithm}
\caption{Assign Noise to Clusters}
\label{algo:assign_noise_to_clusters}
\begin{algorithmic}
\small
\Procedure{AssignNoise}{segments, clusters, n}
    \State $noise \gets \text{segments labeled as noise}$
    
    \While{not converged(clusters)}
        \ForAll{$seg$ in $noise$}
            \State $bestDistance \gets \infty$
            \ForAll{$cl$ in clusters}
                \State $closeSegs \gets n \text{-nearest in } cl \text{ to } seg$
                \State $dist \gets \text{avg distance from } seg \text{ to } closeSegs$
                \If{$dist < bestDist$}
                    \State $bestDistance \gets dist$
                    \State $bestCluster \gets cl$
                \EndIf
            \EndFor
            \State Assign $seg$ to $bestCluster$
        \EndFor
    \EndWhile

    \State \Return clusters
\EndProcedure
\end{algorithmic}
\end{algorithm}

\subsection{Behavior classification and Policy improvement}

Applying the previous steps in the methodology, all line segments are now assigned to a cluster. We take the line segments in the reduced latent space back into state space. Since the state space in this environment is two-dimensional, the clusters are easy to visualize. If the state space was high-dimensional, we could, for visualization purposes, use \ac*{pacmap} to reduce the data down to two dimensions.

Our hypothesis is now that within these clusters, the agent performs a specific type of behavior. Using domain knowledge, in the next section, we attempt to classify the behavior type that occurs within some of the clusters. 

Under our analysis of the behavior within the clusters, we discover inconsistencies in the policy's behavior modes in some regions of the state space. Using domain knowledge, we suggest improvements to the policy's behavior within these regions of the state space. We then test these improvements by selecting actions using either the original policy or a hand-crafted policy within these regions.

\section{Results and Discussion}\label{sec:results_discussion}

In this section, we first show the difference between trajectory clustering directly in latent space and the reduced space produced by \ac*{pacmap} using our methodology. For this, we gathered the results using the MC Policy. Secondly, we investigate the behavior of the MC Policy within a specific area of the state space. That specific area is particularly interesting as it marks the boundary between having or not having sufficient mechanical energy to reach the goal directly. In that region of the state space, the margins between good behavior and bad behavior are small since a single mistake can be the difference between either getting up the goal hill directly or going back to the other hill. Finally, we suggest and implement simple improvements to the MC Policy based on its behavior when the initial state of an episode is within the specific area of the state space just described. 

To create a more precise visualization of the clustered trajectories, we remove whole episodes with trajectories in state space similar to other episodes. This allows us to highlight the general trends in the discovered clusters. Still, when the number of clusters is high, it is not easy to visualize them clearly in a plot. We distinguish between clusters in a single plot using one color for each cluster. Because of this, when there are many clusters, we split a single clustering into two plots. Also, to select color palettes that are as color-blind friendly and distinct as possible, we use the Python package \emph{distinctipy} \cite{distinctipy}.

\subsection{Clustering in latent space vs reduced latent space}

\begin{figure}
    \centering
    \includegraphics{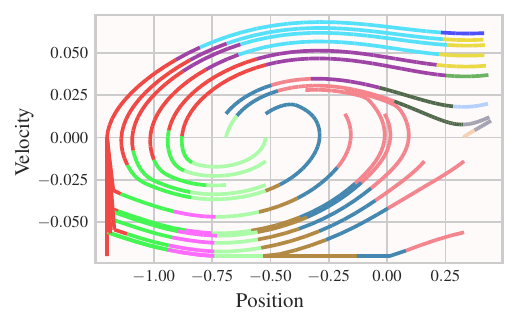}
    \caption{Results from clustering the MC Policy in non-reduced latent space. All lines represent trajectories, where each point in each trajectory corresponds to one time step in an episode of the agent operating in the environment. All trajectories end at the goal car position, which is when position $>= 0.45$, as seen in the upper-right of the plot.}
    \label{fig:policy_C_latent_space_normal}
\end{figure}

Some reasons for performing trajectory clustering in a reduced latent space instead of the entire latent space were highlighted in \cref*{subsec:dim_reduction}. However, to better illustrate the advantage of clustering in the reduced latent space, we show the following results. 

\Cref*{fig:policy_C_latent_space_normal} shows the resulting clusters when clustering directly in latent space. Clustering in this version of the space gives 17 different clusters. The distinctions between the clusters are clear in this plot. However, there seem to be different behavior modes within each cluster. For example, the cluster defined by the blue-green color \fcolorbox{black}{bluegreen}{\rule{0pt}{6pt}\rule{6pt}{0pt}}, going diagonally up to the left from around Pos $ = 0.0$, Vel $ = -0.07$. This cluster seems to contain at least three types of behavior: accelerating toward the right while the velocity is towards the right, accelerating towards the left while the velocity is towards the right, and accelerating towards the left while the velocity is towards the left. For many other clusters, there are also different types of behavior modes, such as the cluster visualized by the red-pink color \fcolorbox{black}{redpink}{\rule{0pt}{6pt}\rule{6pt}{0pt}}, which has similar problems as the blue-green cluster just discussed.

\begin{figure}
    \centering
    \includegraphics{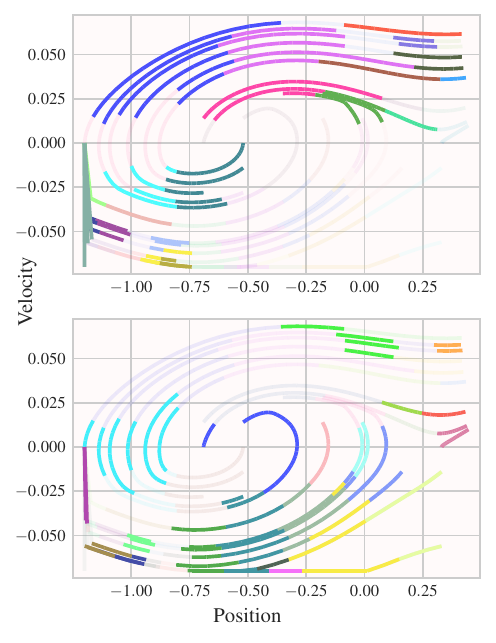}
    \caption{Results from clustering the MC Policy in the reduced latent space. Because of the high number of clusters, this result is shown in two plots for easier visualization.}
    \label{fig:policy_C_two_subplots_normal}
\end{figure}

\Cref*{fig:policy_C_two_subplots_normal} show the clusters generated when clustering in the reduced latent space. There is a much more fine-grained clustering here, and the number of clusters is 42, over twice as many. Many of the boundaries between the clusters in \Cref*{fig:policy_C_latent_space_normal} are still present in this new clustering. At a glance, there seems to still be occurrences of multiple behavior modes in some of the clusters, for instance, for the cluster defined by the dark-blue color \fcolorbox{black}{darkblue}{\rule{0pt}{6pt}\rule{6pt}{0pt}} in the middle of the lower plot in \cref*{fig:policy_C_two_subplots_normal}. However, as we will explain in \cref*{subsec:results_policy_c_clustering}, this is a result of the policy having subpar behavior in this region of the state space. In \cref*{subsec:results_policy_c_improvement}, the behavior within this area is improved with a simple change to the policy.

We hypothesize, from this brief comparison between clustering in the reduced and non-reduced latent space, that \ac*{pacmap} is playing an essential role in the proposed methodology by structuring the latent space data into a format more suitable for trajectory clustering with TRACLUS. However, further work and investigations into why this structuring is suitable for TRACLUS will need to be done.

\subsection{Behavioral clustering of the MC Policy}\label{subsec:results_policy_c_clustering}

\begin{figure}
    \centering
    \includegraphics{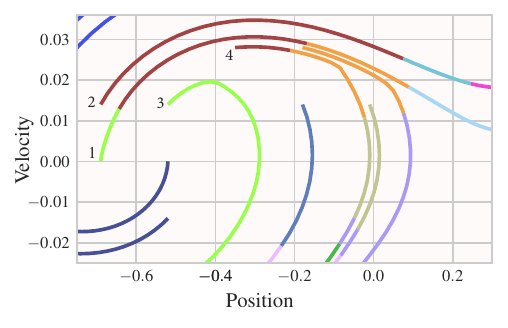}
    \caption{Zoomed in on the area of the state space where the boundary between having enough mechanical energy to reach the goal directly and not having enough mechanical energy exists. Numbers are next to the start of each trajectory/episode we want to discuss for ease of referring to them.}
    \label{fig:policy_C_zoomed_1_subplots}
\end{figure}

We are now looking at the behavior modes of the MC Policy discovered using the proposed methodology within the specific region of the state space described above. \Cref*{fig:policy_C_zoomed_1_subplots} shows a plot of the agent's state space trajectories and assigned clusters within this region. As stated in \cref*{subsec:meth_training_policies}, we operate under the assumption that the purpose of the policy is to make the car reach the goal as quickly as possible. For the four episodes whose behavior modes we want to discuss, we have added numbers next to where the episodes start to refer to them in our discussion. We refer to these episodes as T1-T4 (trajectory 1 - trajectory 4).

The first cluster considered is defined by the rust color \fcolorbox{black}{rust}{\rule{0pt}{6pt}\rule{6pt}{0pt}} which T2 and T4 start in. For this cluster, we classify the behavior as one where the car accelerates toward the right as part of a maneuver to get up the hill to the goal. We suggest that in this behavior mode, the agent infers it has enough mechanical energy to move up the hill. One may then wonder why T4, after moving to the cluster defined by the orange color \fcolorbox{black}{orange}{\rule{0pt}{6pt}\rule{6pt}{0pt}}, accelerates towards the left instead, decreasing the velocity and starting to move to the left. In the next section, we provide results suggesting that this inconsistency indicates the policy has misunderstood the environment in the region of the state space defined by the rust color in T4. 

We now discuss the behavior within the cluster defined by the yellow-green color \fcolorbox{black}{yellowgreen}{\rule{0pt}{6pt}\rule{6pt}{0pt}} which T1 and T3 start in. We classify the behavior within this cluster as a transition behavior, where the agent switches the direction in which it accelerates. T1 and T3 each have very different results, with T1 moving straight up the right hill to the goal and T3 charging up more mechanical energy by moving up the left hill before going to the goal. We hypothesize that the policy has also misunderstood the environment in this region of the state space and that, at least for T1, this initial switching behavior leads to poorer results. This claim is demonstrated in the next section by altering the policy in the region around the yellow-green clustering in T1. 

\subsection{Improvement of the MC Policy based on behavior clustering}\label{subsec:results_policy_c_improvement}

\begin{figure}
    \centering
    \includegraphics{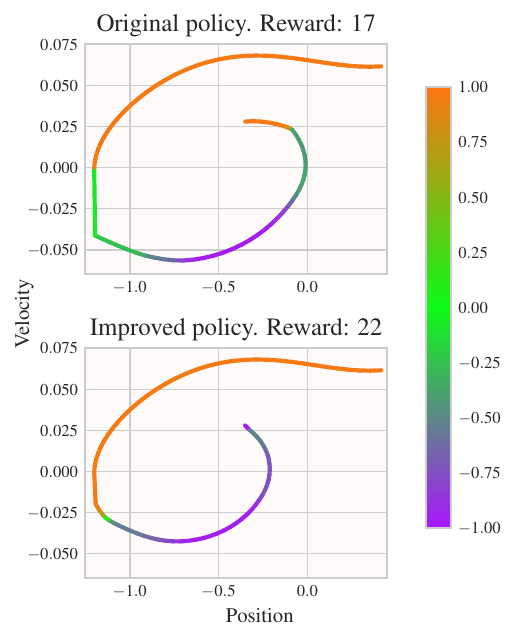}
    \caption{The first plot shows running the MC Policy in the environment from the initial state $s_0 = [-0.35, 0.028]$. The second plot shows the same, with the only change being that we force the first action to be $a_0 = -1$.}\label{fig:changed_policy_035_0028}
\end{figure}

Now, we have two hypotheses for regions of the state space where the policy has misunderstood the environment's dynamics and is, as a result, performing sub-par. The first region is around the rust-colored \fcolorbox{black}{rust}{\rule{0pt}{6pt}\rule{6pt}{0pt}} cluster at the beginning of T4 in \cref*{fig:policy_C_zoomed_1_subplots}. We hypothesize that since the agent does not go straight up the right hill to the goal, as we suggest the agent is attempting in this behavior mode, another type of behavior could be better in this region of the state space. We suggest accelerating to the left instead of the right as the agent currently does. We highlight the difference between these two initial types of behavior in \cref*{fig:changed_policy_035_0028}. This plot shows that if the policy just chose to accelerate to the left a single time step at the start, the accumulated reward through the episode would be 22 instead of 17. 

\begin{figure}
    \centering
    \includegraphics{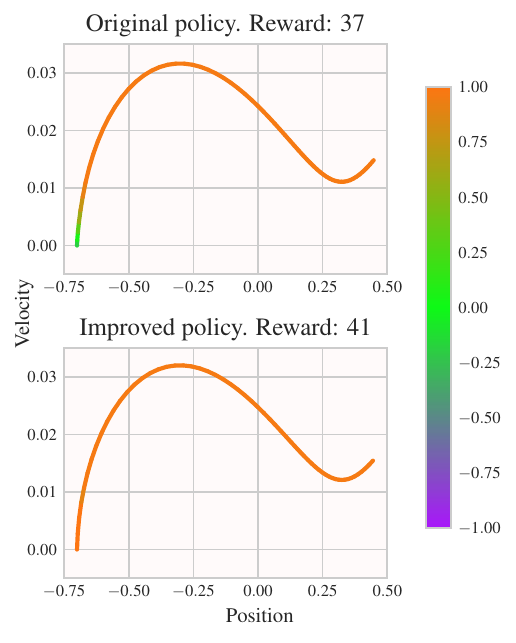}
    \caption{The first plot shows running the MC Policy in the environment from the initial state $s_0 = [-0.7, 0.0]$. The second plot shows the same, with the only change being that we force the three first actions to be $a_0 = a_1 = a_2 = 1.0$.}
    \label{fig:changed_policy_075_00}
\end{figure}

The second region is around the yellow-green \fcolorbox{black}{yellowgreen}{\rule{0pt}{6pt}\rule{6pt}{0pt}} cluster at the beginning of T1. We hypothesize that the switching behavior occurring within this region is sub-par, and the agent would be better off if it accelerates towards the right from the start. A comparison between the policy's behavior and the proposed alternative behavior within the yellow-green cluster can be seen in \cref*{fig:changed_policy_075_00}. With the proposed change, the agent would receive a cumulative reward over the episode of 41 instead of 37. 





\section{Conclusion}\label{sec:conclusion}

We propose a novel methodology for discovering behavior modes of a \ac*{drl} agent by clustering trajectories in the latent space of the agent's neural network policy. These behavior modes can reveal areas where the policy does not produce suitable control actions. By employing dimensionality reduction via \ac*{pacmap} and trajectory clustering with TRACLUS, we could observe and analyze the behavioral clusters of a policy solving the classic control task of Mountain Car. 

We analyzed the state space region where the car chooses between a direct uphill attempt or building up mechanical energy and found slight misinterpretations of the environment by the policy, leading to suboptimal performance. Focusing on this critical region, we made some simple improvements to the policy. These adjustments were based on the identified behavioral clusters and resulted in increased rewards during the episodes considered. Our modifications increase the reward from 17 to 22 in one scenario and 37 to 41 in another. This demonstrates the practical applications of our clustering-based analysis.

For a simple environment like Mountain Car, detecting the sub-par behavior in the policy could have been done by examining the agent's behavior using domain knowledge alone. However, the proposed methodology gives a convenient way to supplement the domain knowledge with a concise description of the agent's policy. Future work will examine more complicated and higher-dimensional \ac*{drl} environments. 

Future work will refine the greedy method for noise segment reassignment presented in this paper. Addressing policy and environment stochasticity and uncertainty is also crucial. In addition, developing interactive visualizations is essential for applications requiring human-machine interfaces.

To conclude, although there are some challenges and limitations to the approach described in this paper, this initial examination of the proposed methodology has provided promising insights and demonstrated tangible improvements in policy performance for the Mountain Car control task.

\section*{Acknowledgment}
The Research Council of Norway supported this work through the EXAIGON project, project number 304843.

\bibliographystyle{IEEEtran}
\bibliography{mylib}

\begin{thebibliography}{10}
\providecommand{\url}[1]{#1}
\csname url@samestyle\endcsname
\providecommand{\newblock}{\relax}
\providecommand{\bibinfo}[2]{#2}
\providecommand{\BIBentrySTDinterwordspacing}{\spaceskip=0pt\relax}
\providecommand{\BIBentryALTinterwordstretchfactor}{4}
\providecommand{\BIBentryALTinterwordspacing}{\spaceskip=\fontdimen2\font plus
\BIBentryALTinterwordstretchfactor\fontdimen3\font minus
  \fontdimen4\font\relax}
\providecommand{\BIBforeignlanguage}[2]{{%
\expandafter\ifx\csname l@#1\endcsname\relax
\typeout{** WARNING: IEEEtran.bst: No hyphenation pattern has been}%
\typeout{** loaded for the language `#1'. Using the pattern for}%
\typeout{** the default language instead.}%
\else
\language=\csname l@#1\endcsname
\fi
#2}}
\providecommand{\BIBdecl}{\relax}
\BIBdecl

\bibitem{badia2020agent57}
A.~P. Badia, B.~Piot, S.~Kapturowski, P.~Sprechmann, A.~Vitvitskyi, Z.~D. Guo,
  and C.~Blundell, ``Agent57: Outperforming the atari human benchmark,'' in
  \emph{International conference on machine learning}.\hskip 1em plus 0.5em
  minus 0.4em\relax PMLR, 2020, pp. 507--517.

\bibitem{brunke2022safe}
L.~Brunke, M.~Greeff, A.~W. Hall, Z.~Yuan, S.~Zhou, J.~Panerati, and A.~P.
  Schoellig, ``Safe learning in robotics: From learning-based control to safe
  reinforcement learning,'' \emph{Annual Review of Control, Robotics, and
  Autonomous Systems}, vol.~5, pp. 411--444, 2022.

\bibitem{wang2021understanding}
\BIBentryALTinterwordspacing
Y.~Wang, H.~Huang, C.~Rudin, and Y.~Shaposhnik, ``Understanding how dimension
  reduction tools work: An empirical approach to deciphering t-sne, umap,
  trimap, and pacmap for data visualization,'' \emph{Journal of Machine
  Learning Research}, vol.~22, no. 201, pp. 1--73, 2021. [Online]. Available:
  \url{http://jmlr.org/papers/v22/20-1061.html}
\BIBentrySTDinterwordspacing

\bibitem{lee2007trajectory}
J.-G. Lee, J.~Han, and K.-Y. Whang, ``Trajectory clustering: a
  partition-and-group framework,'' in \emph{Proceedings of the 2007 ACM SIGMOD
  international conference on Management of data}, 2007, pp. 593--604.

\bibitem{mannor2004dynamic}
S.~Mannor, I.~Menache, A.~Hoze, and U.~Klein, ``Dynamic abstraction in
  reinforcement learning via clustering,'' in \emph{Proceedings of the
  twenty-first international conference on Machine learning}, 2004, p.~71.

\bibitem{sutton1999between}
R.~S. Sutton, D.~Precup, and S.~Singh, ``Between mdps and semi-mdps: A
  framework for temporal abstraction in reinforcement learning,''
  \emph{Artificial intelligence}, vol. 112, no. 1-2, pp. 181--211, 1999.

\bibitem{watkins1992q}
C.~J. Watkins and P.~Dayan, ``Q-learning,'' \emph{Machine learning}, vol.~8,
  pp. 279--292, 1992.

\bibitem{mukherjee2019clustergan}
S.~Mukherjee, H.~Asnani, E.~Lin, and S.~Kannan, ``Clustergan: Latent space
  clustering in generative adversarial networks,'' in \emph{Proceedings of the
  AAAI conference on artificial intelligence}, vol.~33, no.~01, 2019, pp.
  4610--4617.

\bibitem{gjaerum2023model}
V.~B. Gj{\ae}rum, I.~Str{\"u}mke, J.~L{\o}ver, T.~Miller, and A.~M. Lekkas,
  ``Model tree methods for explaining deep reinforcement learning agents in
  real-time robotic applications,'' \emph{Neurocomputing}, vol. 515, pp.
  133--144, 2023.

\bibitem{guo2021edge}
W.~Guo, X.~Wu, U.~Khan, and X.~Xing, ``Edge: Explaining deep reinforcement
  learning policies,'' \emph{Advances in Neural Information Processing
  Systems}, vol.~34, pp. 12\,222--12\,236, 2021.

\bibitem{he2021explainable}
L.~He, N.~Aouf, and B.~Song, ``Explainable deep reinforcement learning for uav
  autonomous path planning,'' \emph{Aerospace science and technology}, vol.
  118, p. 107052, 2021.

\bibitem{lin2014visualising}
K.~W.~E. Lin, H.~Anderson, N.~Agus, C.~So, and S.~Lui, ``Visualising singing
  style under common musical events using pitch-dynamics trajectories and
  modified traclus clustering,'' in \emph{2014 13th International Conference on
  Machine Learning and Applications}.\hskip 1em plus 0.5em minus 0.4em\relax
  IEEE, 2014, pp. 237--242.

\bibitem{mustafa2021gtraclus}
H.~Mustafa, C.~Barrus, E.~Leal, and L.~Gruenwald, ``Gtraclus: a local
  trajectory clustering algorithm for gpus,'' in \emph{2021 IEEE 37th
  International Conference on Data Engineering Workshops (ICDEW)}.\hskip 1em
  plus 0.5em minus 0.4em\relax IEEE, 2021, pp. 30--35.

\bibitem{ester1996density}
M.~Ester, H.-P. Kriegel, J.~Sander, X.~Xu \emph{et~al.}, ``A density-based
  algorithm for discovering clusters in large spatial databases with noise,''
  in \emph{kdd}, vol.~96, no.~34, 1996, pp. 226--231.

\bibitem{chen2003noisy}
J.~Chen, M.~K. Leung, and Y.~Gao, ``Noisy logo recognition using line segment
  hausdorff distance,'' \emph{Pattern recognition}, vol.~36, no.~4, pp.
  943--955, 2003.

\bibitem{towers_gymnasium_2023}
\BIBentryALTinterwordspacing
M.~Towers, J.~K. Terry, A.~Kwiatkowski, J.~U. Balis, G.~d. Cola, T.~Deleu,
  M.~Goulão, A.~Kallinteris, A.~KG, M.~Krimmel, R.~Perez-Vicente, A.~Pierré,
  S.~Schulhoff, J.~J. Tai, A.~T.~J. Shen, and O.~G. Younis, ``Gymnasium,'' Mar.
  2023. [Online]. Available: \url{https://zenodo.org/record/8127025}
\BIBentrySTDinterwordspacing

\bibitem{stable-baselines3}
\BIBentryALTinterwordspacing
A.~Raffin, A.~Hill, A.~Gleave, A.~Kanervisto, M.~Ernestus, and N.~Dormann,
  ``Stable-baselines3: Reliable reinforcement learning implementations,''
  \emph{Journal of Machine Learning Research}, vol.~22, no. 268, pp. 1--8,
  2021. [Online]. Available: \url{http://jmlr.org/papers/v22/20-1364.html}
\BIBentrySTDinterwordspacing

\bibitem{haarnoja2018soft}
T.~Haarnoja, A.~Zhou, K.~Hartikainen, G.~Tucker, S.~Ha, J.~Tan, V.~Kumar,
  H.~Zhu, A.~Gupta, P.~Abbeel \emph{et~al.}, ``Soft actor-critic algorithms and
  applications,'' \emph{arXiv preprint arXiv:1812.05905}, 2018.

\bibitem{optuna_2019}
T.~Akiba, S.~Sano, T.~Yanase, T.~Ohta, and M.~Koyama, ``Optuna: A
  next-generation hyperparameter optimization framework,'' in \emph{Proceedings
  of the 25th {ACM} {SIGKDD} International Conference on Knowledge Discovery
  and Data Mining}, 2019.

\bibitem{mackiewicz1993principal}
A.~Ma{\'c}kiewicz and W.~Ratajczak, ``Principal components analysis (pca),''
  \emph{Computers \& Geosciences}, vol.~19, no.~3, pp. 303--342, 1993.

\bibitem{BellmanDynamicProgramming}
R.~E. Bellman, \emph{Dynamic Programming}.\hskip 1em plus 0.5em minus
  0.4em\relax Princeton University Press, 1957.

\bibitem{beyer1999nearest}
K.~Beyer, J.~Goldstein, R.~Ramakrishnan, and U.~Shaft, ``When is “nearest
  neighbor” meaningful?'' in \emph{Database Theory—ICDT’99: 7th
  International Conference Jerusalem, Israel, January 10--12, 1999 Proceedings
  7}.\hskip 1em plus 0.5em minus 0.4em\relax Springer, 1999, pp. 217--235.

\bibitem{distinctipy}
\BIBentryALTinterwordspacing
J.~Roberts, J.~Crall, K.-M. Ang, and Y.~Brandt,
  ``alan-turing-institute/distinctipy: v1.2.3,'' Sep. 2023. [Online].
  Available: \url{https://doi.org/10.5281/zenodo.8355862}
\BIBentrySTDinterwordspacing

\end{thebibliography}

\end{document}